\documentclass[conference]{IEEEtran}
\IEEEoverridecommandlockouts
\usepackage{cite}
\usepackage{amsmath,amssymb,amsfonts}
\usepackage{algorithmic}
\usepackage{graphicx}
\usepackage{textcomp}
\usepackage{balance}
\usepackage{xcolor}
\usepackage{siunitx}
\usepackage{caption}
\usepackage{subcaption}
\usepackage{multirow}
\usepackage{makecell}

\DeclareMathOperator*{\argmin}{arg\,min}

\def\BibTeX{{\rm B\kern-.05em{\sc i\kern-.025em b}\kern-.08em
    T\kern-.1667em\lower.7ex\hbox{E}\kern-.125emX}}
\begin{document}

\title{Generative AI Empowered LiDAR Point Cloud Generation with Multimodal Transformer
\vspace{-1ex}
}

\author{\vspace{0ex}
\normalsize
Mohammad Farzanullah\IEEEauthorrefmark{1}, Han Zhang\IEEEauthorrefmark{1}, Akram Bin Sediq\IEEEauthorrefmark{2}, Ali Afana\IEEEauthorrefmark{2} and Melike Erol-Kantarci\IEEEauthorrefmark{1}\\
\vspace{0ex}
\IEEEauthorblockA{\IEEEauthorrefmark{1}
    School of Electrical Engineering and Computer Science, University of Ottawa, Ottawa, ON, Canada} 
\IEEEauthorblockA{\IEEEauthorrefmark{2}
    Ericsson Inc., Ottawa, ON, Canada}
Emails: \{mfarz086, hzhan363, melike.erolkantarci\}@uottawa.ca
\\ \{akram.bin.sediq, ali.afana\}@ericsson.com
\vspace{-3ex}
} 

\maketitle

\begin{abstract}
Integrated sensing and communications is a key enabler for the 6G wireless communication systems. The multiple sensing modalities will allow the base station to have a more accurate representation of the environment, leading to context-aware communications. Some widely equipped sensors such as cameras and RADAR sensors can provide some environmental perceptions. However, they are not enough to generate precise environmental representations, especially in adverse weather conditions. On the other hand, the LiDAR sensors provide more accurate representations, however, their widespread adoption is hindered by their high cost.
%This paper proposes a novel approach to enhance the wireless communication systems by synthesizing LiDAR point clouds from images and RADAR data using a multimodal transformer architecture.
%\notered{Separate it into two sentences here: This paper proposes a novel approach to enhance the wireless communication systems by synthesizing LiDAR point clouds from images and RADAR data. Specifically, it uses a multimodal transformer architecture ... (generalize the novelty of the work, for example, you can say it uses a multimodal transformer architecture and pre-trained foundation models to enable an accurate lidar data generation). Or you can say it uses a multimodal transformer architecture to learn latent representations of different modalities.}
This paper proposes a novel approach to enhance the wireless communication systems by synthesizing LiDAR point clouds from images and RADAR data. Specifically, it uses a multimodal transformer architecture and pre-trained encoding models to enable an accurate LiDAR generation.
%Our proposed algorithm employs pre-trained vision transformers and depth encoders to extract representations from camera images and RADAR readings, thereby reducing training costs and improving efficiency. 
%We evaluate our algorithm on the DeepSense 6G dataset, specifically curated for wireless applications 
The proposed framework is evaluated on the DeepSense 6G dataset, which is a real-world dataset curated for context-aware wireless applications.
Our results demonstrate the efficacy of the proposed approach in accurately generating LiDAR point clouds. We achieve a modified mean squared error of 10.3931. Visual examination of the images indicates that our model can successfully capture the majority of structures present in the LiDAR point cloud for diverse environments. This will enable the base stations to achieve more precise environmental sensing. By integrating LiDAR synthesis with existing sensing modalities, our method can enhance the performance of various wireless applications, including beam and blockage prediction. 
%Overall, this paper underscores the potential of multimodal transformer architectures in advancing the capabilities of future wireless communication systems.
\end{abstract}


\section{Introduction} \label{Section:Introduction}

Artificial Intelligence (AI) is a rapidly evolving field that encompasses various techniques and approaches aimed to enable machines to mimic human intelligence and cognitive abilities. At its core, AI involves the development of algorithms and models that enable machines to process data, learn from it, and make decisions or predictions accordingly. Recently, AI algorithms have been widely utilized for the optimization of the 5th Generation (5G) and 6th Generation (6G) wireless communication systems \cite{qiao2021survey}.

Future 6G systems will combine higher frequency bands from millimeter wave (mmWave) to THz, wider bandwidth, and denser antenna arrays to integrate signal sensing and communication, mutually enhancing their capabilities \cite{isac}. The communication systems can help to improve sensing capabilities. 
Simultaneously, the base stations are equipped with diverse modalities such as visual, sensing, and localization capabilities. This configuration paves the way for context-aware communications, which is beneficial for applications such as beam prediction, handover control, and reduced overhead for tracking channel state information.
%\notered{Change it like this: Simultaneously, the base stations are equipped with visual capabilities and localization capabilities giving the deployment of sensors for multiple sensing modalities. It paves the way for context-aware communications, which is beneficial for applications such as beam prediction, handover control, ... Can you rewrite this more clearly?}, paving way for context-aware communications, such as more accurate beam prediction, efficient handover, and reduced overhead for tracking channel state information.

Some existing works used visual input and RADAR input to perform wireless communication-related tasks \cite{demirhan2022radar, charan2022vision}.  However, cameras can be less reliable in harsh weather conditions, such as snow and fog, and RADAR sensors only produce a sparser representation of the environment. Measurements generated by these sensors in sub-optimal sensing environments may not be solely relied upon for context-aware communications. In contrast, LiDAR sensors can provide a more accurate and stable representation of the environment. However, LiDAR sensors have not yet been widely deployed at the base stations considering their high cost \cite{singh2023depth}. 
%Even though LiDAR sensors have been mostly used for autonomous driving applications 
Based on their exceptional achievements demonstrated on autonomous driving applications, LiDAR sensors are believed to present a more promising sensing solution for mmWave base stations. 
In the existing studies, the LiDAR data has proven effective for various wireless applications, such as beam prediction \cite{jiang2022lidar, tian2023multimodal} and handover efficiency \cite{wu2022lidar}. Therefore, there is a need to synthesize LiDAR data with other modalities to enhance the wireless communication performance without introducing extra costs.

%\notered{Please add some description about the generation capabilities of transformers or multi-modal transformers here. You can start with this: On the other hand, transformer-based generative models have recently attracted much attention because of their outstanding capabilities ... multi-modal data. Therefore, this architecture can be suitable ...}

On the other hand, transformer-based generative models have recently attracted much attention due to their remarkable ability to generate coherent and contextually relevant text \cite{vaswani2017attention}. 
These models, built upon the transformer architecture, have been applied across various domains, including Natural Language Processing (NLP) and computer vision.
%Transformers  are a popular deep learning architecture that has been used for many applications, such as Natural Language Processing (NLP), and computer vision. 
The transformer architecture relies on the concept of self-attention mechanisms, which allows the model to weigh the importance of different words in a sequence. This self-attention mechanism enables transformers to capture long-range dependencies in the input data. 
A multimodal transformer encoder \cite{xu2023multimodal} refers to a variation or extension of the transformer architecture specifically designed to process multimodal data. Transformers typically utilize self-attention mechanisms, allowing each token to attend to other tokens within the same input sequence. In a multimodal setting, cross-modal attention mechanisms are introduced to allow tokens from one modality to attend to tokens from other modalities, enabling the model to learn inter-modal relationships.

Inspired by these thoughts, in this paper, we utilize a multimodal transformer architecture to generate LiDAR point clouds from images and RADAR data.  
%We evaluate our model on the DeepSense 6G dataset. Our results demonstrate that we can accurately predict LiDAR point clouds for diverse environments, allowing the base station to have more accurate sensing of the environment. 
%\notered{You don't need to have a new subsection for contributions and you don't need to write "In this paper ..." again. Also, you only need to mention the performance once. You can Please merge these together. It should be: ... for autonomous vehicles. The major contributions of ... We evaluate our model on the DeepSense 6G dataset. Our results demonstrate that we can accurately predict LiDAR point clouds for diverse environments, allowing the base station to have more accurate sensing of the environment.}
%\subsection{Our contributions}
%In this paper, we propose multimodal transformer-based generative AI algorithms to predict three-dimensional (3D) LiDAR point clouds based on the inputs of camera and RADAR sensing information. 
%\notered{You can remove this part and merge it with the end of the introduction. 1. In this paper, we propose .... 2. The main contributions are .... 3. Our proposed framework shows a good performance... 4. The subsequent sections of the paper are organized as follows:}
The major contributions of our paper can be summarized as follows:
\begin{itemize}
    \item We designed a novel multimodal transformer-based LiDAR synthesis/generation algorithm, comprising image encoder, RADAR encoder, depth encoder, multimodal transformer, and LiDAR decoder. 
    %\notered{You can remove from compromising}. 
    The inputs to the deep learning algorithm are the camera image and RADAR sensor data. To the best of our knowledge, we are the first to use multimodal transformer to learn the relationship between different input modalities to predict LiDAR.
    The synthesized LiDAR data can help in the decision-making process for many wireless applications such as beam prediction and blockage prediction, and for autonomous vehicles.
    \item Most existing LiDAR generation works use an encoder-decoder architecture, and train the complete model. Instead, we use a pre-trained vision transformer and depth encoder during the encoding stage to learn the image and RADAR representation, and this effectively reduces the training cost.
    %\item We have evaluated the performance of our proposed algorithm on the DeepSense 6G dataset, a real-world multimodal dataset specifically designed for wireless applications.  
\end{itemize}

We evaluate our model on the DeepSense 6G dataset. Our results demonstrate that we can accurately predict LiDAR point clouds for diverse environments, allowing the base station to have more accurate sensing of the environment.
The visual inspection of our generated LiDAR images show that our model was able to learn majority of the structures captured by the actual LiDAR sensors.
%Our proposed framework shows a good performance in generating LiDAR sensor information from images and RADAR data. 

The subsequent sections of the paper are organized as follows: Section \ref{Section:RelatedWorks} offers a review of relevant literature. Next, Section \ref{Section:Methodology} discusses in detail the data preprocessing and our Machine Learning model. This is followed by the dataset and training details in Section \ref{Section:SimulationSettings} and results in Section \ref{Section:Results}. Finally, we conclude our paper in Section \ref{Section:Conclusions}.

\section{Related work} \label{Section:RelatedWorks}

A few studies have attempted to predict LiDAR point clouds from other modalities. The authors in \cite{balemans2021predicting} used a convolutional stacked autoencoder to predict two-dimensional (2D) LiDAR data from a series of ultrasonic images in an indoor environment. The ultrasonic images yield a sparser representation of the environment, making them insufficient as a sensing modality. An encoder-decoder architecture was used, with the aim of learning distance for each angle.
Another study aimed to learn the depth map from images and RADAR point clouds for a self-driving car dataset \cite{singh2023depth}. A two-stage architecture was developed, where quasi-dense depth was learned in the first stage, and an encoder-decoder architecture was used in the second stage to generate a depth map. 
The depth map was generated using scaffolding technique on LiDAR data.
The authors in \cite{vacek2021learning} sought to learn the LiDAR reflective intensity based on camera and depth map using a U-Net architecture. 
However, \cite{balemans2021predicting} used single modality at the input, while \cite{singh2023depth, vacek2021learning} used an encoder-decoder architecture without benefiting from the inter-modal relationships. %\notered{In contrast, our paper ... (in one sentence)}
%\notered{Explain how our work is different from existing works. You shorten the explanation of other works a little bit if you don't have enough space for the paper.}
In contrast, to the best of our knowledge, we are the first to use multimodal transformer-based generative AI model to predict the 3D LiDAR data from camera and RADAR data, with pre-trained encoding stage. Furthermore, \cite{balemans2021predicting, singh2023depth, vacek2021learning} evaluated their models on self-driving vehicles dataset, while we use the DeepSense 6G dataset, which was collected for wireless communications scenario.

The recent popularity of generative AI algorithms inspired the research community for the densification of the LiDAR point clouds. 
Densification refers to the process of increasing the point density within a given area, resulting in a more detailed representation of the environment.
A U-Net architecture was used for the generation and densification of LiDAR point clouds in \cite{zyrianov2022learning}. Similarly, \cite{nakashima2023lidar} used probabilistic diffusion models and \cite{xiong2023learning} used a variation of encoder-decoder architecture for the densification of LiDAR point clouds. 
However, these works fall in the category of synthetic data generation or densification, as they do not use other modalities to predict the LiDAR point cloud. 
%\notered{when we compare our results with the groundtruth and they do the same, can we say our performance is better. We are saying we are doing something different but not showing if it's better. Please leave this comment and reply in blue. we'll delete those comments later.} \noteblue{I did not find any work that is doing exactly the same thing (predicting LiDAR from RADAR and image). In 5,11,12, they use different modalities, while in 13-15, they are densifying LiDAR using genAI models.} \notered{can you clarify better in the above summary. Because I thought they were similar. Maybe you should talk about densifying and how is it different. not too long though, just a short statement} \noteblue{Added a sentence explaining densification. The difference with our work is mentioned in the last sentence "they do not use other modalities to predict the LiDAR point cloud. "}

%The authors in \cite{hess2024lidarclip} developed LidarCLIP, a model that learns embedding for LiDAR point clouds corresponding to images in CLIP embedding space. 
%\notered{Still, explain how our work is different from existing works.}

%\notered{I would still suggest making related work a separate section. You can move the applications of lidar data from the introduction to the related works, and you need to compare existing works with our work to show the novelty. So it will be longer than now and make a separate section. In this way, the introduction can be connected with the contributions and can have better consistency.}

\section{Methodology} \label{Section:Methodology}

In this section, we discuss the model architecture of our proposed multimodal transformer-based LiDAR point cloud generation framework. 
%We use camera images and RADAR sensor information for the generation of LiDAR point clouds. 
The inputs to the framework are the camera images and RADAR sensor information, and the expected output is the generated LiDAR image.

As discussed in Section \ref{Section:Introduction}, the LiDAR sensor is expensive to install and might not be feasible to install at every base station. This calls for a strategy to generate LiDAR data from other modalities. %\notered{Maybe no need for a new paragraph for the next paragraph?}
In this paper, we aim to generate LiDAR depth data based on camera images and RADAR data. Mathematically, the goal is to find the optimal model parameters $W^*$ that minimize a given loss, which can be formulated as follows:
\begin{align}
    W^* = \argmin_W L(Y, f(X;W))
\end{align}
where $W$ are the parameters of our model, $X$ are the inputs (camera and RADAR sensor information), $Y$ is the actual LiDAR sensor information, and $f()$ is the model we use for prediction. $f(X;W)$ denotes the LiDAR point cloud predicted by our model. $L()$ is the loss function, which will be defined in Section \ref{Section:SimulationSettings}.

\begin{figure*} [t]
    \centering
    \includegraphics[width=0.97\linewidth]{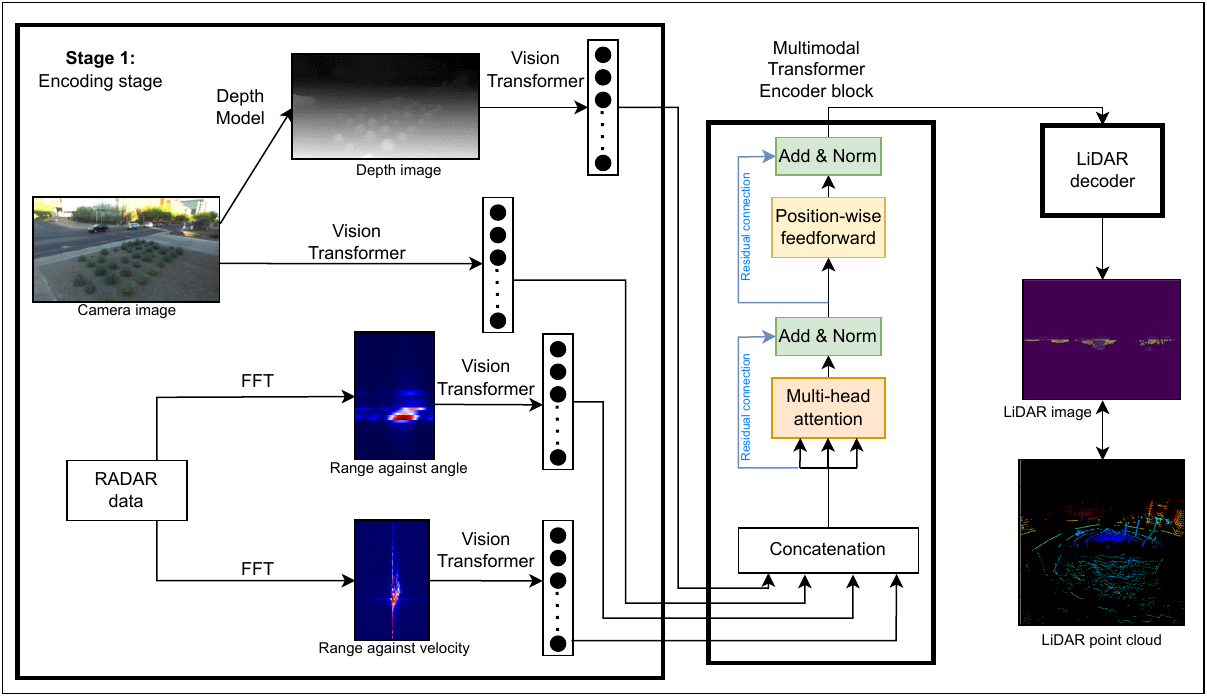}
    \caption{Our model architecture. The input modalities are passed through vision transformers to produce embeddings. The embeddings are passed through the transformer encoder, followed by LiDAR decoder to generate the final LiDAR image.}
    \label{fig:modelarch}
    \vspace{-3ex}
\end{figure*}

The Fig. \ref{fig:modelarch} shows our model architecture. %\notered{There are three main stages in this architecture, namely ...} 
There are three main stages in this architecture, namely the encoding stage, the multimodal transformer stage, and the LiDAR decoder stage. It is worth noting that the encoding stage uses pre-trained models. Only the transformer encoder stage and the LiDAR decoder need to be trained. 

%During the encoding stage, all the modalities are transformed into embeddings of size 768 \notered{is this 768 part of your hyperparameter setting? if it is, can you just say embedings of the same size? If it is something related to vision transformer, can you explain a little bit} using the vision transformer.
During the encoding stage, the pre-trained vision transformer converts all the modalities into embeddings of size 768, which represents a fixed output dimensionality inherent to the vision transformer architecture.
%The camera image is passed\notered{you can replace passed with other words like preprocessed, processed, converted into, Or you can change the order, for examples, the embedings are then obtained through .. } through the vision transformer to produce camera embeddings. 
The camera image is converted to camera embeddings using the vision transformer.
Simultaneously, the camera image is also passed through a pre-trained depth model to generate a relative depth image. 
The depth image provides a richer contextual understanding of an image, enabling more accurate scene understanding.
The depth image is passed through a vision transformer to produce depth embeddings.
%\notered{Add a few explanations here why you use the depth model. For example, it can grab semantic meanings from the image and convert it to depth information, which is useful for lidar prediction. Something like this. Please rephrase that.} 
Similarly, the RADAR range against angle matrix and range against velocity matrix are fed through the vision transformer to produce another pair of embeddings. These four embeddings, each of size 768, are passed to the transformer encoder block. The transformer encoder model produces a latent space vector. This is followed by a LiDAR decoder that produces the final LiDAR image.
The details of each of these parts are discussed in the remaining part of this section.

\subsection{Modailty Preprocessing}

\subsubsection{RADAR preprocessing}
In the dataset provided in \cite{DeepSense}, a frequency modulated continuous wave RADAR is used for the collection of data. The data consists of 3D complex I/Q RADAR measurements of the shape (number of receive antenna $\times$ samples per chirp $\times$ chirps per frame). The resolution of the RADAR is 60 cm. 

%The RADAR preprocessing technique is used as in \cite{demirhan2022radar}\notered{My suggestions is you mention this paper in related works but don't mention it here. We can directly start with how we process the data}. 
First, the range Fast Fourier Transform (FFT) is obtained by performing FFT in the second dimension, which corresponds to samples per chirp in the time domain. The result of the range FFT is then subjected to an extra FFT operation along the first dimension, which indicates the number of receive antennas, in order to obtain the range against angle matrix. Lastly, by applying FFT to the range FFT output along the third dimension, which is indicative of chirps per frame, the range against velocity matrix is obtained.
%We pass the input raw RADAR data to range, velocity, and angle Fast Fourier Transform (FFT) to produce two new matrices. The two matrices show the range against velocity and range against angle. 
These two matrices are depicted in the form of an image in Fig. \ref{fig:modelarch}.
%\notered{Do you have math formulations here? If not just ignore this suggestion.}.

\subsubsection{LiDAR preprocessing}
%\notered{Can you explain a little bit here why you need a lidar preprocessing. Is this because it is difficult to train the model directly with 3D points data? Or this is just because we need a better presentation of our result? please explain it here.}
The LiDAR data is provided in the form of 3D point clouds. The 3D point cloud consists of multiple points in the 3D space that reflect the light. It is not feasible to directly learn the LiDAR point cloud as it encompasses the entire 3D space. Recognizing the computational challenges posed by the vastness of LiDAR point cloud data, an alternative strategy is pursued to facilitate effective learning.

We convert each point in the LiDAR point cloud from Cartesian coordinates to the polar coordinates. This forms a 2D matrix, where the dimensions of the matrix represent the reference angle from the XY plane (termed as $\theta$), and the reference angle from the Z-axis (termed as $\phi$). This can be viewed as a 2D LiDAR image, where the dimensions of the matrix represent $\theta$ and $\phi$, and the values in the matrix represent the distance from the origin. For $\theta$, a granularity of $0.25^{\circ}$ is used for the whole range from $-180^{\circ}$ to $180^{\circ}$. Meanwhile, for $\phi$, a granularity of $0.25^{\circ}$ is used for the range from $-60^{\circ}$ to $-5^{\circ}$ and from $5^{\circ}$ to $62^{\circ}$, and a granularity of $0.015625^{\circ}$ is used for the range from  $-5^{\circ}$ to $5^{\circ}$. A finer granularity is used for $-5^{\circ}$ to $5^{\circ}$ because most of the objects of interest (cars, people) lie in this range. This resulted in a target matrix of size $1440 \times 1088$. As shown in Fig. \ref{fig:modelarch}, the 3D LiDAR point cloud and 2D LiDAR image are reciprocal (one can be generated from the other).

In this way, our problem can be converted into predicting the distance from the origin for all the possible angles in space.

\subsection{Modality encoding stage}

The modality encoding stage comprises pre-trained models only.

%\subsubsection{Depth Encoder} \notered{To reduce the difficulty of model training}We use the “Depth Anything” model \cite{yang2024depth} to \notered{give the model more intuitive depth information in addition to the original image input (Please rephrase it in a better way)} find the relative depth of an image taken from a base station. This is an encoder-decoder model, using the DINOv2 encoder and the DPT \notered{explain DPT here} decoder. 

\subsubsection{Depth Model} To reduce the difficulty of model training, we use the pre-trained “Depth Anything” model \cite{yang2024depth} to find the relative depth of an image taken from a base station. The relative depth enriches our model with a greater contextual understanding of the surroundings. Relative depth integration improves the model's comprehension of spatial relationships and object locations. This is an encoder-decoder model, using the DINOv2 encoder and the dense prediction transformer (DPT) decoder. 

%\subsubsection{Vision Transformer} \notered{In this framework, we use vision transformer (cite) to convert pre-trained multi-modal information into embeddings of the same size. For data of each modality, a separate vision transformer is used. The vision transformer first ... Please adjust the order of this subsection and start with this} The Vision Transformer \cite{dosovitskiy2020image} uses the Transformer Encoder architecture to convert an image into an embedding of size 768. Each image is partitioned into fixed-size patches. The patches are linearly embedded, position embeddings are added, and the resulting vector is passed through a transformer encoder. 
\subsubsection{Vision Transformer} In this framework, we use pre-trained vision transformer \cite{dosovitskiy2020image} to convert each multimodal information into embeddings of size 768. 
 The vision transformer first partitions each image into fixed-size patches. The patches are linearly embedded, position embeddings are added, and the resulting vector is passed through a transformer encoder, to produce the final embeddings. 
 For data of each modality, a separate vision transformer is used. The vision transformer acts as an encoder for all the modalities, lowering the dimension of each modality to 768.

\subsection{Multimodal Transformer Encoder}

In this framework, a multimodal transformer encoder is also used to capture long-range dependencies in the multimodal input data. The multimodal transformer cross-modal attention mechanisms are introduced to allow tokens from one modality to attend to tokens from other modalities, enabling the model to learn inter-modal relationships.

In the proposed LiDAR generation framework, the embeddings of all of the four modalities are first concatenated and then fed into the multimodal transformer encoder. The multihead attention is able to compute multiple attention mechanisms in parallel, which enables the model to focus on different parts of the input sequence simultaneously. The position-wise feedforward layer in transformers enables localized non-linear transformations at each position within the input sequence, capturing relevant patterns in the input. The output is a latent space embedding of size 1024.

\subsection{LiDAR decoder}
\begin{figure} [h]
    \centering
    \vspace{-1ex}
    \includegraphics[width=\linewidth]{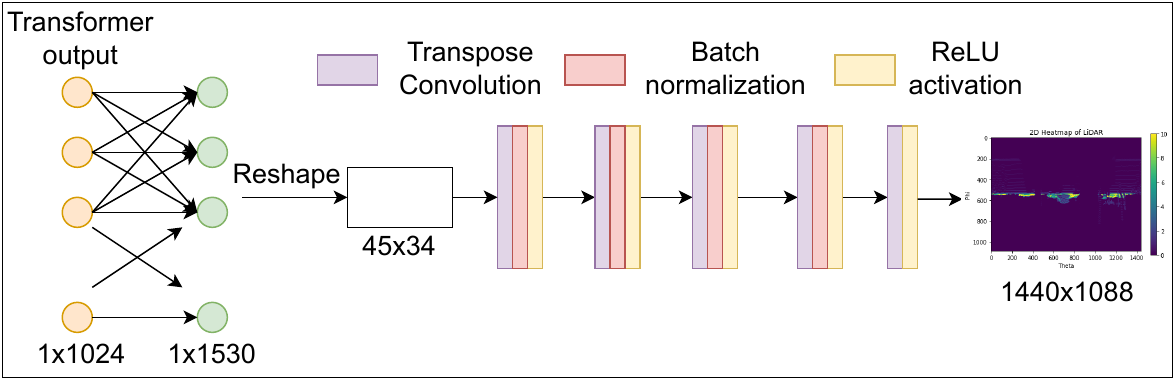}
    \caption{The LiDAR decoder consists of 5 transpose convolutional layers, that upscale the image to our LiDAR image size.}
    \label{fig:lidarDecoder}
    \vspace{-1ex}
\end{figure}

The last part of the proposed framework is the LiDAR decoder. In this stage, the embeddings produced by the multimodal transformer encoder are fed into the LiDAR decoder for 2D LiDAR image generation.
The LiDAR decoder consists of a fully connected layer followed by a series of transpose convolutional layers to upscale the output at each stage. Each layer $i$ consists of a convolutional transpose layer with $l_i$ filters.
Filters in Convolutional Neural Networks act as feature detectors, convolving over input data to extract relevant patterns, edges, and textures, enabling the network to learn hierarchical representations in an image.
We use a kernel size of 4, with a stride of 2 and padding of 1. This guarantees that the image undergoes a doubling in size with each layer. Each convolutional transpose layer is followed by batch normalization and rectified linear unit (ReLU) activation function. 

%\notered{can you explain why there is a filter and what it is for in maybe one sentence?}.
%\notered{Is there any needs that we define this $\mathcal{L}$ and $\{l_1, l_2, l_3, l_4\}$ here? I mean this is just a definition and we don't use it in this section. You can define it later in the simulation part before you use them.}

\section{Simulation settings} \label{Section:SimulationSettings}

\subsection{Dataset}

We use Scenarios 31-34 of the DeepSense 6G dataset \cite{DeepSense}, where the data for each scenario was collected at a different location. The data for scenario 31 and 32 was collected during the day, while the data for scenario 33 and 34 was collected during the night. The dataset consists of an outdoor environment. 
%where the base station is equipped with camera, RADAR, and LiDAR sensors. 
The base station is equipped with different sensors so data of different modalities are provided in the dataset, including the imaging modality and the RADAR modality. The ground-truth LiDAR data is also provided for training and testing. Other than the above mentioned data, the 64-element receiving power vector is also provided for beam prediction. The primary goal of the dataset was for sensing aided beam prediction, where the sensing data can be utilized to improve beam management. 
With this dataset, we generate LiDAR data with camera images and RADAR data. By comparing the synthetic LiDAR data with real LiDAR data, we can evaluate the effectiveness of our proposed LiDAR generation framework.

\subsection{Implementation Details}
For each scenario, we split the data into 60\% (11199 data samples), 20\% (3733 data samples), and 20\% (3733 data samples) for training, validation, and test set respectively.
For each scenario, the training data comprises the first 60\% of the samples, followed by 20\% for validation, and the remaining 20\% for testing, aimed at minimizing correlation between training and testing sets.
%Furthermore, the training data consists of the first 60\% of the samples in time, the validation data consists of the next 20\% samples, while test data consists of the last 20\% of the samples. This was done to reduce correlation between the training and the testing set. \notered{Can you rephrase this more concisely, for example, three datasets are split over different consecutive periods to reduce the correlation between the training and the testing set.}

The LiDAR image is sparse, with varying density in different regions of the image, particularly concentrated around values of $\phi$ close to 0.
Training the model using the simple mean squared error results in the model over-fitting, causing it to predict close to 0 for the entire sparse LiDAR image. 
To counter this, we propose modified mean squared error (MMSE) loss, where the denser areas of the LiDAR image were given 10 times more weight than sparser regions.
For an image with $M$ pixels, the MMSE loss is defined as:
\begin{align}
\frac{1}{M} \sum_{i=1}^{M} \alpha_i (y_i - \hat{y}_i)^2 ,       \begin{cases}
        \alpha_i = 10, & \phi \in [-1.71875, 2.1875] \\
        \alpha_i = 1,  & \text{otherwise}
    \end{cases} \label{eq:loss}
\end{align}
where $\alpha_i$ refers to weight given to pixel $i$ in the image, $y_i$ refers to the actual LiDAR image, and $\hat{y}_i$ refers to the LiDAR image predicted by our model. The values of $\alpha$ were obtained through trial-and-error. 
%The idea was to prioritize the model to learn the denser region.  \notered{Can you explain this before the MMSE definition? First, explain that the density of lidar information varies. So we give different weights to the denser regions and sparser regions ... The MMSE is defined as follows ... }
\begin{table*} [ht!]
        \vspace{1ex}
	\caption{MMSE comparison between different model complexities and all-zeroes prediction.}	
	\centering %
	\begin{tabular}{| c| c| c| c| c| c|}				
		\hline
         & \multicolumn{4}{c|}{\textbf{Scenario}} & \\ \hline
		$\mathcal{L}$ & \textbf{31} & \textbf{32} & \textbf{33} & \textbf{34} & \textbf{Overall} \\ \hline
        \{32, 16, 8, 8\} & 18.8390 & 16.8927 & 17.9924 & 22.5443 & 19.2058 \\ \hline
        \{64, 32, 16, 16\} & 12.3774 & 12.6932 & 12.8715 & 18.4655 & 13.9846 \\ \hline
        \{128, 64, 32, 32\} & 11.0890 & 10.4354 & 10.3220 & 14.4871 & 11.6153 \\ \hline
        \{256, 128, 64, 64\} & 10.4239 & 9.3329 & 9.2480 & 12.1860 & \textbf{10.3931} \\ \hline
        \{512, 256, 128, 128\} & 11.1808 & 9.1056 & 9.1524 & 12.4217 & 10.6743 \\ \hline
        Predict all zeroes & 48.3617 & 26.7759 & 30.1806 & 30.8811 & 38.5810 \\ \hline
        \{256, 128, 64, 64\} - without transformer encoder & \makecell{12.2389 \\ (+14.83\%)}  & \makecell{9.2966 \\ (-0.39\%)} & \makecell{9.6717 \\ (+4.38\%)} & \makecell{11.9774 \\ (-1.74\%)} & \makecell{11.1118 \\ (+6.47\%)} \\ \hline
 \end{tabular}
 \vspace{-4ex}
	\label{Table:losses}
\end{table*}	
We used an Adam Optimizer with a batch size of 32. The training was performed for 20 epochs, where the learning rate for the first 10 epochs was 0.001 and was reduced to 0.0001 for the last 10 epochs. 

We used one transformer encoder layer with 12 heads in the multi-head attention model. The dimension of the feedforward network model was set to 2048, and the dropout was set to 0.1. The output of the transformer encoder layer had size of 1024. 

For the LiDAR decoder, we first had a fully connected layer with 1350 neurons. The 1350 nodes were then reshaped to a matrix of 45x30. Next, we use 5 convolutional transpose layers, each consisting of convolutional transpose filters, batch normalization, and ReLU activation function. Let $\mathcal{L}$ denote the set consisting of convolutional filters in the first four layers, i.e., $\{l_1, l_2, l_3, l_4\}$. In the final layer, we use 1 filter as it produces the final LiDAR image.
In our experiments, we varied the number of filters, and obtained the best results with {256, 128, 64, 64} filters in the first four layers. 
The weights for the convolutional layers were initialized using a normal distribution with a mean of 0 and a standard deviation of 0.02, while batch normalization parameters were initialized with a mean of 1 and standard deviation 0.02.

PyTorch library was used to develop the model. The model was trained using the NVIDIA A100 Tensor core GPU. It took around 3 hours to train 20 epochs.

\section{Results and Discussions} \label{Section:Results}

In this section, we display our experimental results.

\begin{figure} [ht]
    \centering
    \vspace{-3ex}
    \includegraphics[width=\linewidth]{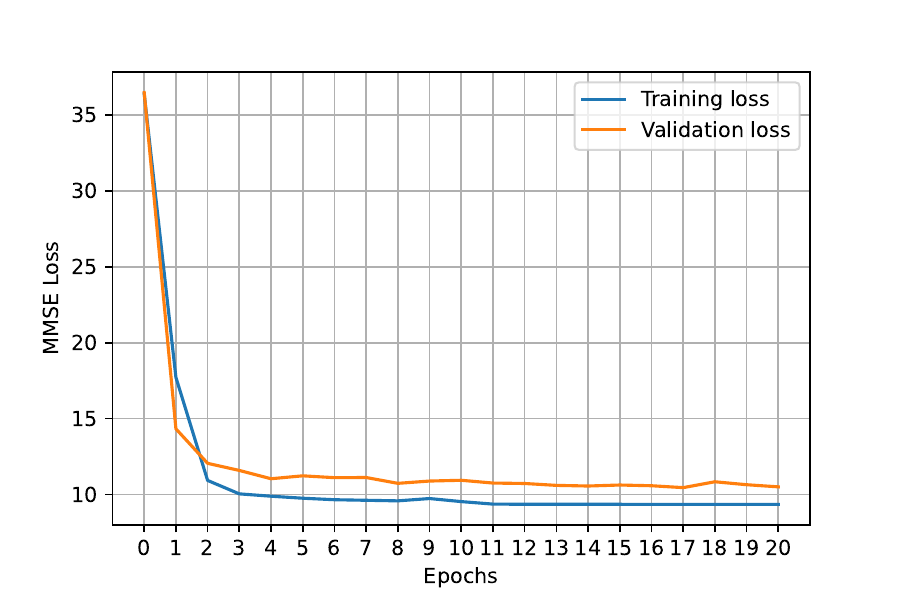}
    \caption{Training and validation loss against number of epochs.}
    \label{fig:losses}
    \vspace{-1ex}
\end{figure}

Fig. \ref{fig:losses} shows the training set and validation set loss against the number of epochs during the training phase. Before the start of the training, the training and validation loss is larger than 35. By the end of the training, the training and validation MMSE losses have reduced by 74.3\% and 71.2\% to 9.3 and 10.5, respectively. Moreover, it can be observed that reducing the learning rate from $0.001$ to $0.0001$ after 10 epochs slightly helps in reducing MMSE further.

\renewcommand{\cellalign}{cc}
\renewcommand{\theadalign}{cc}

In Table \ref{Table:losses}, we show the performance of the model when we vary the number of filters in the convolutional transpose layers in the LiDAR decoder. Furthermore, as the LiDAR image is sparse and contains the value of 0 for more than 95\% of the pixels, we develop a lower benchmark called all-zeroes prediction, where the output LiDAR image consists of all zeroes. 
We provide another benchmark, where we train the model using the LiDAR decoder that results in the least MMSE, without the transformer encoder block (the modality embeddings are directly passed to the LiDAR decoder).
%We do not provide an upper benchmark, as no existing work is sufficiently close to our work. 

The predict all-zeroes benchmark results in overall MMSE of 38.5810 for the testing set. In comparison, all of our models perform better. At first, increasing the number of filters in the LiDAR decoder helps in reducing the MMSE loss. The best results are obtained with \{256, 128, 64, 64\} filters in the convolutional layers, resulting in an overall testing loss of 10.3931. Increasing the complexity of the LiDAR decoder further does not help in reducing the overall MMSE loss. Moreover, the benchmark without transformer encoder achieves an overall MMSE of 11.1118, which is higher than that of the proposed method by 6.47\%.
%\notered{can you rewrite this. use - existing transformer models do this or dont do this, etc. the same with the following. dont say no transformer...}\noteblue{rewritten the sentence.}
The scenario 34 obtains MMSE of 12.1860, which is higher than other scenarios.  The model without transformer encoder achieves almost the same MMSE as our proposed method for scenario 34. One potential explanation might be that the data was gathered during nighttime, leading to a decline in image clarity. 
%\notered{check the results section again.You use different tenses in different paragraphs. I prefer present tense not the past tense.}\noteblue{changed the tense to present.}
Interestingly, the highest MMSE in the all-zeroes prediction is observed for scenario 31. However, our model exhibits superior performance in scenario 31, achieving a notably low MMSE of 10.4239. In comparison, the model without transformer encoder achieves MMSE of 12.2389, which is 14.83\% higher as compared to our proposed method.

\begin{figure} [t]
\centering
    \begin{subfigure}{.49\textwidth}
      \centering
      \includegraphics[width=1\linewidth]{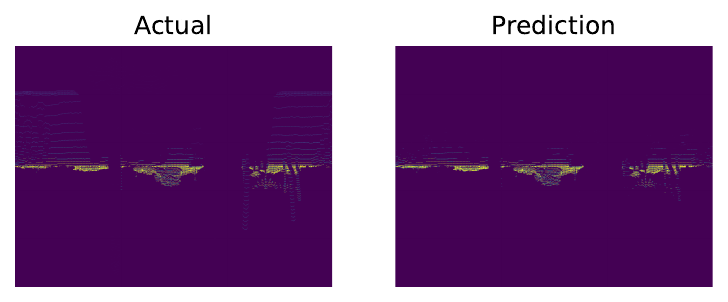}
      \vspace{-4ex}
      \caption{Scenario 31}
      \label{fig:output31}
    \end{subfigure}%
    
    \begin{subfigure}{.49\textwidth}
      \centering
      \includegraphics[width=1\linewidth]{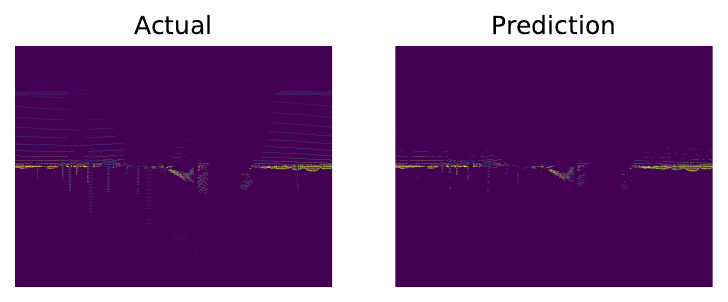}
      \vspace{-4ex}
      \caption{Scenario 32}
      \label{fig:output32}
    \end{subfigure}%

    \begin{subfigure}{.49\textwidth}
      \centering
      \includegraphics[width=1\linewidth]{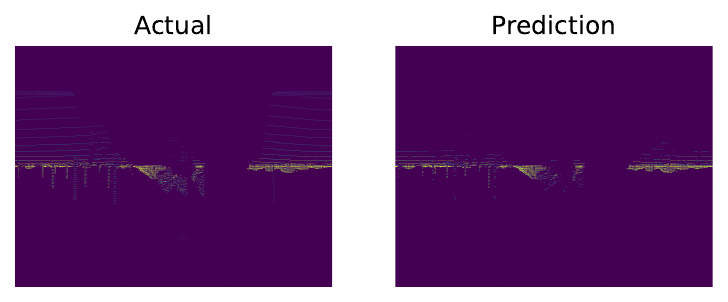}
      \vspace{-4ex}
      \caption{Scenario 33}
      \label{fig:output33}
    \end{subfigure}%

    \begin{subfigure}{.49\textwidth}
      \centering
      \includegraphics[width=1\linewidth]{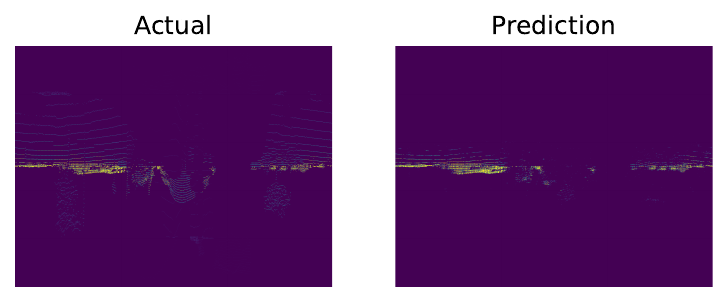}
      \vspace{-4ex}
      \caption{Scenario 34}
      \label{fig:output34}
    \end{subfigure}%

\caption{Comparison between actual and predicted LiDAR for a random LiDAR point cloud from each scenario.}
\vspace{-4ex}
\label{fig:VisualOutputs}
\end{figure}

Fig. \ref{fig:VisualOutputs} compares the actual and predicted LiDAR image. We randomly sample a single LiDAR image from each scenario, and visually compare the actual LiDAR image and the LiDAR image predicted by our best model (with \{256, 128, 64, 64\} filters in LiDAR decoder). For each scenario, the model is able to learn most of the structures captured by the LiDAR image. As discussed earlier, the loss is highest in Scenario 34, which is also apparent by visual inspection of Fig. \ref{fig:output34}. However, overall the model is able to accurately learn most of the structures captured by the LiDAR sensor for all four scenarios. This shows the robustness of our model for multiple diverse environments.

\section{Conclusions and future work} \label{Section:Conclusions}

%\notered{Please put the second sentence as the first sentence: LiDAR sensor is useful for context-aware communications, and it can help optimize many wireless communication tasks such as blockage prediction and beam prediction. In this work,}
The LiDAR sensor proves valuable for context-aware communications, offering optimization benefits across various wireless communication tasks like handover prediction and beam prediction.
In this work, we develop a multimodal transformer encoder based LiDAR point cloud generation model. Our model allows the generation of LiDAR sensor information without installing an expensive LiDAR sensor. The inputs to our model are the camera image and RADAR sensing information. We use pre-trained models to encode input modalities. The multimodal transformer model facilitates learning the relationship between modalities. Finally, we use a convolutional neural network based LiDAR decoder to generate a LiDAR image. Our experiments show that we were able to obtain MMSE of 10.3931. The visual inspection of the images shows that our model is able to learn most of the structures within the LiDAR image for diverse scenarios.

In the future, we can utilize the synthesized LiDAR point cloud to enhance various wireless communication tasks such as beam and blockage prediction. Furthermore, our work can be useful for other industries as well, like the self-driving cars.

\vspace{0ex}
%+++++++++++++++++++++++++++++++++++++++++++++++++++++++++++++++++++++++++++++++++++++++++++++++
\balance
\bibliographystyle{IEEEtran} 
\bibliography{globecom_paper}
\balance \balance

%+++++++++++++++++++++++++++++++++++++++++++++++++++++++++++++++++++++++++++++++++++++++++++++++
\end{document}